\documentclass[11pt,a4paper]{article}
\usepackage[hyperref]{acl2020}

\usepackage{microtype}

\aclfinalcopy 

\usepackage{setspace}
\usepackage[]{algorithm2e}
\usepackage{times}
\usepackage{latexsym}
\usepackage{amsmath}
\usepackage{multirow}
\usepackage{graphicx}
\usepackage{amssymb}
\usepackage{tabularx}
\usepackage{arydshln}
\usepackage{amsfonts}
\usepackage{todonotes}
\usepackage{fancyvrb}
\usepackage{tikz}
\usepackage{xcolor}
\usepackage{color}
\usepackage{float}
\usepackage{natbib}
\usepackage{dcolumn}
\usepackage[english]{babel}
\usepackage{balance}
\usepackage[bottom]{footmisc}
\usepackage{capt-of}
\usepackage{fixltx2e}
\usepackage{flushend}

\usepackage{pgfplots}
\usepackage{pgfplotstable} 
\pgfplotsset{compat=newest}

\title{A Relational Memory-based Embedding Model for Triple Classification and Search Personalization}

\author{Dai Quoc Nguyen${}^{1}$, Tu Dinh Nguyen${}^{2}$, Dinh Phung${}^{1}$\\
${}^{1}$Monash University, Australia\\
${}^{2}$Trusting Social\\
${}^{1}${\tt{{\{dai.nguyen,dinh.phung\}@monash.edu}}}\\
${}^{2}${\tt{tu@trustingsocial.com}}
}

\begin{document}
\maketitle

\begin{abstract}
Knowledge graph embedding methods often suffer from a limitation of memorizing valid triples to predict new ones for triple classification and search personalization problems. To this end, we introduce a novel embedding model, named R-MeN, that explores a relational memory network to encode potential dependencies in relationship triples. R-MeN considers each triple as a sequence of 3 input vectors that recurrently interact with a memory using a transformer self-attention mechanism. Thus R-MeN encodes new information from interactions between the memory and each input vector to return a corresponding vector. Consequently, R-MeN feeds these 3 returned vectors to a convolutional neural network-based decoder to produce a scalar score for the triple. Experimental results show that our proposed R-MeN obtains state-of-the-art results on SEARCH17 for the search personalization task, and on WN11 and FB13 for the triple classification task. 

\end{abstract}

\section{Introduction}
Knowledge graphs (KGs) -- representing the genuine relationships among entities in the form of triples \textit{(subject, relation, object)} denoted as \textit{(s, r, o)} -- are often insufficient for knowledge presentation due to the lack of many valid triples \citep{West:2014}. 
Therefore, research work has been focusing on inferring whether a new triple missed in KGs is likely valid or not \citep{bordes2011learning,NIPS2013_5071,NIPS2013_5028}.
As summarized in \citep{NickelMTG15,Nguyen2017}, KG embedding models aim to compute a score for each triple, such that valid triples have higher scores than invalid ones.

Early embedding models such as TransE \citep{NIPS2013_5071}, TransH \citep{AAAI148531}, TransR \citep{AAAI159571}, TransD \citep{ji-EtAl:2015:ACL-IJCNLP}, DISTMULT \citep{Yang2015} and ComplEx \citep{Trouillon2016} often employ simple linear operators such as addition, subtraction and multiplication.
Recent embedding models such as ConvE \citep{Dettmers2017} and CapsE \citep{Nguyen2019CapsE} successfully apply deep neural networks to score the triples.

Existing embedding models are showing promising performances {mainly} for knowledge graph completion, where the goal is to infer a missing entity given a relation and another entity.
But in real applications, less mentioned, such as triple classification \citep{NIPS2013_5028} that aims to predict whether a given triple is valid, and search personalization \citep{vu2017search} that aims to re-rank the relevant documents returned by a user-oriented search system given a query, these models do not effectively capture potential dependencies among entities and relations from existing triples to predict new triples.

To this end, we leverage on the relational memory network \citep{santoro2018relational} to propose R-MeN to infer a valid fact of new triples.
In particular, R-MeN transforms each triple along with adding positional embeddings into a sequence of 3 input vectors.
R-MeN then uses a transformer self-attention mechanism \citep{vaswani2017attention} to guide the memory to interact with each input vector to produce an encoded vector.  
As a result, R-MeN feeds these 3 encoded vectors to a convolutional neural network (CNN)-based decoder to return a score for the triple.
In summary, our main contributions are as follows:

\begin{itemize}
    
\item We present R-MeN -- a novel KG embedding model to memorize and encode the potential dependencies among relations and entities for two real applications of triple classification and search personalization.

\item Experimental results show that R-MeN obtains better performance than up-to-date embedding models, in which R-MeN produces new state-of-the-art results on SEARCH17 for the search personalization task, and a new highest accuracy on WN11 and the second-highest accuracy on FB13 for the triple classification task.

\end{itemize}

\section{The proposed R-MeN}
\label{sec:ourmodel}

\begin{figure}[!ht]
\centering
\includegraphics[width=0.45\textwidth]{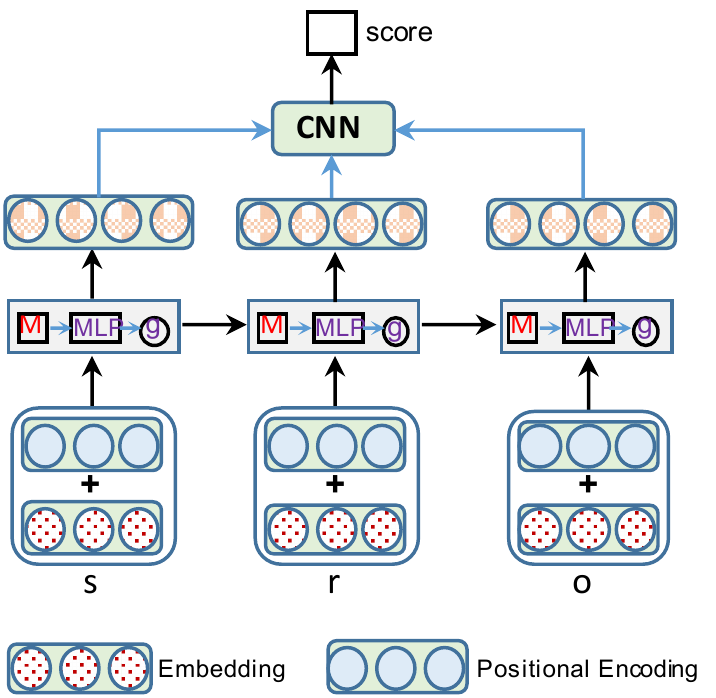}
\caption{Processes in our proposed R-MeN for an illustration purpose. ``M'' denotes a memory. ``MLP'' denotes a multi-layer perceptron. ``g'' denotes a memory gating. ``CNN'' denotes a convolutional neural network-based decoder.}
\label{fig:rmen}
\end{figure}

Let $\mathcal{G}$ be a KG database of valid triples in the form of \textit{(subject, relation, object)} denoted as \textit{(s, r, o)}. 
KG embedding models aim to compute a score for each triple, such that valid triples obtain higher scores than invalid triples.

We denote $\boldsymbol{\mathsf{v}}_s$, $\boldsymbol{\mathsf{v}}_r$ and $\boldsymbol{\mathsf{v}}_o \in \mathbb{R}^{d}$ as the embeddings of $s$, $r$ and $o$, respectively.
Besides, we hypothesize that relative positions among $s$, $r$ and $o$ are useful to reason instinct relationships; hence we add to each position a positional embedding.
Given a triple \textit{(s, r, o)}, we obtain a sequence of 3 vectors $\{\boldsymbol{\mathsf{x}}_1, \boldsymbol{\mathsf{x}}_2, \boldsymbol{\mathsf{x}}_3\}$  as:
\begin{eqnarray}
\boldsymbol{\mathsf{x}}_1 &=& \textbf{W}\left(\boldsymbol{\mathsf{v}}_s + \boldsymbol{\mathsf{p}}_1\right) + \textbf{b} \nonumber\\ \boldsymbol{\mathsf{x}}_2 &=& \textbf{W}\left(\boldsymbol{\mathsf{v}}_r + \boldsymbol{\mathsf{p}}_2\right)  + \textbf{b} \nonumber\\
\boldsymbol{\mathsf{x}}_3 &=& \textbf{W}\left(\boldsymbol{\mathsf{v}}_o + \boldsymbol{\mathsf{p}}_3\right)  + \textbf{b} \nonumber
\end{eqnarray}
\noindent where $\textbf{W} \in \mathbb{R}^{k\times d}$ is a weight matrix, and  $\boldsymbol{\mathsf{p}}_1, \boldsymbol{\mathsf{p}}_2$ and $\boldsymbol{\mathsf{p}}_3 \in \mathbb{R}^{d}$ are positional embeddings, and $k$ is the memory size.

We assume we have a memory $M$ consisting of $N$ rows wherein each row is a memory slot.
We use $M^{(t)}$ to denote the memory at timestep $t$, and $M^{(t)}_{i,:}  \in \mathbb{R}^{k}$ to denote the $i$-th memory slot at timestep $t$. 
We follow \citet{santoro2018relational} to take $\boldsymbol{\mathsf{x}}_t$ to update $M^{(t)}_{i,:}$ using the multi-head self-attention mechanism \citep{vaswani2017attention} as:
\begin{eqnarray}
& \hat{M}^{(t+1)}_{i,:} & = [\hat{M}^{(t+1),1}_{i,:}\oplus\hat{M}^{(t+1),2}_{i,:}\oplus \nonumber \\
&& \ \ \ \ \  ... \oplus \hat{M}^{(t+1),H}_{i,:}] \nonumber\\
& \mathsf{with} \ \ \hat{M}^{(t+1),h}_{i,:} & = \alpha_{i,N+1,h}\left(\textbf{W}^{h,V}\boldsymbol{\mathsf{x}}_t\right) \nonumber\\
&& \ \ \ \ \  + \sum_{j=1}^N\alpha_{i,j,h}\left(\textbf{W}^{h,V}M^{(t)}_{j,:}\right) \nonumber
\end{eqnarray}
\noindent  where $H$ is the number of attention heads, and $\oplus$ denotes a vector concatenation operation.
Regarding the $h$-th head, $\textbf{W}^{h,V} \in \mathbb{R}^{n\times k}$ is a value-projection matrix, in which $n$ is the head size and $k=nH$. Note that $\left\{\alpha_{i,j,h}\right\}_{j=1}^N$ and $\alpha_{i,N+1,h}$ are attention weights, which are computed using the $\mathsf{softmax}$ function over scaled dot products as:
\begin{eqnarray}
\alpha_{i,j,h} &=& \frac{\exp\left(\beta_{i,j,h}\right)}{\sum_{m=1}^{N+1}\exp\left(\beta_{i,m,h}\right)} \nonumber\\
\alpha_{i,N+1,h} &=& \frac{\exp\left(\beta_{i,N+1,h}\right)}{\sum_{m=1}^{N+1}\exp\left(\beta_{i,m,h}\right)} \nonumber\\
\mathsf{with} \ \ \beta_{i,j,h} &=& \frac{\left(\textbf{W}^{h,Q}M^{(t)}_{i,:}\right)^\mathsf{T}\left(\textbf{W}^{h,K}M^{(t)}_{j,:}\right)}{\sqrt{n}} \nonumber\\
\beta_{i,N+1,h} &=& \frac{\left(\textbf{W}^{h,Q}M^{(t)}_{i,:}\right)^\mathsf{T}\left(\textbf{W}^{h,K}\boldsymbol{\mathsf{x}}_t\right)}{\sqrt{n}} \nonumber
\end{eqnarray}
\noindent where $\textbf{W}^{h,Q} \in \mathbb{R}^{n\times k}$ and $\textbf{W}^{h,K} \in \mathbb{R}^{n\times k}$ are query-projection and key-projection matrices, respectively.
As following \citet{santoro2018relational}, we feed a residual connection between $\boldsymbol{\mathsf{x}}_t$ and $\hat{M}^{(t+1)}_{i,:}$ to a multi-layer perceptron followed by a memory gating to produce an encoded vector $\boldsymbol{\mathsf{y}}_t \in \mathbb{R}^k$ for timestep $t$ and the next memory slot $M^{(t+1)}_{i,:}$ for timestep $(t+1)$.

As a result, we obtain a sequence of 3 encoded vectors $\{\boldsymbol{\mathsf{y}}_1, \boldsymbol{\mathsf{y}}_2, \boldsymbol{\mathsf{y}}_3\}$ for the triple $(s, r, o)$. 
We then use a CNN-based decoder to compute a score for the triple as:
\begin{equation}
f\left(s,r,o\right) = \mathsf{max}\left(\mathsf{ReLU}\left([\boldsymbol{\mathsf{y}}_1, \boldsymbol{\mathsf{y}}_2, \boldsymbol{\mathsf{y}}_3]\ast\bold{\Omega}\right)\right)^\mathsf{T} \bold{w} \nonumber
\end{equation}
where we view $[\boldsymbol{\mathsf{y}}_1, \boldsymbol{\mathsf{y}}_2, \boldsymbol{\mathsf{y}}_3]$ as a matrix in $\mathbb{R}^{k\times3}$; $\bold{\Omega}$ denotes a set of filters in $\mathbb{R}^{m\times3}$, in which $m$ is the window size of filters; $\bold{w} \in \mathbb{R}^{|\bold{\Omega}|}$ is a weight vector; $\ast$ denotes a convolution operator; and $\mathsf{max}$ denotes a max-pooling operator.
Note that we use the max-pooling operator -- instead of the vector concatenation of all feature maps used in ConvKB \citep{Nguyen2018} -- to capture the most important feature from each feature map, and to reduce the number of weight parameters.

We illustrate our proposed R-MeN as shown in Figure \ref{fig:rmen}.
In addition, we employ the Adam optimizer \citep{kingma2014adam} to train R-MeN by minimizing the following loss function \citep{Trouillon2016,Nguyen2018}:

{\small
\begin{align}
\mathcal{L} =  \sum_{\substack{(s, r, o) \in \{\mathcal{G} \cup \mathcal{G}'\}}} \log\left(1 + \exp\left(- t_{(s, r, o)} \cdot f\left(s,r,o\right)\right)\right) \nonumber
 \label{equal:objfunc}
\end{align}
}
  \vspace{-10pt}
\begin{equation*}
\text{in which, } t_{(s, r, o)} = \left\{ 
  \begin{array}{l}
  1\;\text{for } (s, r, o)\in\mathcal{G}\\
 -1\;\text{for } (s, r, o)\in\mathcal{G}'
  \end{array} \right.
\end{equation*}

\noindent where $\mathcal{G}$  and $\mathcal{G}'$  are collections of valid and invalid triples, respectively.  $\mathcal{G}'$ is generated by corrupting valid triples in $\mathcal{G}$. 

\section{Experimental setup}

\subsection{Task description and evaluation}

\subsubsection{Triple classification}
\label{subsec:triplecls}
The triple classification task is to predict whether a given triple $(s, r, o)$ is valid or not \citep{NIPS2013_5028}.
Following \citet{NIPS2013_5028}, we use two benchmark datasets WN11 and FB13, in which each validation or test set consists of the same number of valid and invalid triples.
It is to note in the test set that \citet{NIPS2013_5028} did not include triples that either or both of their subject and object entities also appear in a different relation type or order in the training set, to avoid reversible relation problems.
Table \ref{tab:datasets1} gives statistics of the experimental datasets.

\begin{table}[!ht]
\centering
\resizebox{7.5cm}{!}{
\def\arraystretch{1.1}
\setlength{\tabcolsep}{0.4em}
\begin{tabular}{l|lllll}
\hline
\bf Dataset & \bf $\#\mathcal{E}$ & \bf $\#\mathcal{R}$  & \multicolumn{3}{l}{\bf \#Triples in train/valid/test} \\
\hline
FB13 & 75,043  & 13 &  316,232 &  11,816 & 47,466\\
WN11 &38,696 & 11 & 112,581 & 5,218 & 21,088 \\
\hline
\end{tabular}
}
\caption{Statistics of the experimental datasets. $\bf \#\mathcal{E}$ is the number of entities. $\bf \#\mathcal{R}$ is the number of relations.}
\label{tab:datasets1}
\end{table}

Each relation $r$ has a threshold $\theta_r$ computed by maximizing the micro-averaged classification accuracy on the validation set.
If the score of a given triple $(s, r, o)$ is above $\theta_r$, then this triple is classified as a valid triple, otherwise, it is classified as an invalid one.

\subsubsection{Search personalization}
In search personalization, given a submitted \textit{query} for a \textit{user}, we aim to re-rank the \textit{documents} returned by a search system, so that the more the returned documents are relevant for that query, the higher their ranks are.
We follow \citep{vu2017search,Nguyen2018ConvKBfull,Nguyen2019CapsE} to view a relationship of the submitted query, the user and the returned document as a \textit{(s, r, o)}-like triple \textit{(query, user, document)}.
Therefore, we can adapt our R-MeN for the search personalization task.

We evaluate our R-MeN on the benchmark dataset SEARCH17 \citep{vu2017search} as follows: (i) We train our model and use the trained model to compute a score for each \textit{(query, user, document)} triple. (ii) We sort the scores in the descending order to obtain a new ranked list.
(iii) We employ two standard evaluation metrics: mean reciprocal rank (MRR) and Hits@1.
For each metric, the higher value indicates better ranking performance.

\subsection{Training protocol}
\subsubsection{Triple classification}

We use the common Bernoulli strategy \citep{AAAI148531,AAAI159571} when sampling invalid triples. 
For WN11, we follow \citet{guu2015traversing} to initialize entity and relation embeddings in our R-MeN by averaging word vectors in the relations and entities, i.e., $\boldsymbol{\mathsf{v}}_{american\_arborvitae} = \frac{1}{2}\left(\boldsymbol{\mathsf{v}}_{american} + \boldsymbol{\mathsf{v}}_{arborvitae}\right)$, in which these word vectors are taken from the Glove 50-dimensional pre-trained embeddings \citep{pennington2014glove} (i.e., d = 50).
For FB13, we use entity and relation embeddings produced by TransE to initialize entity and relation embeddings in our R-MeN, for which we obtain the best result for TransE on the FB13 validation set when using $\mathit{l}_2$-norm, learning rate at 0.01, margin $\gamma$ = 2 and d = 50.

Furthermore, on WN11, we provide our new fine-tuned result for TransE using our experimental setting, wherein we use the same initialization taken from the Glove 50-dimensional pre-trained embeddings to initialize entity and relation embeddings in TransE.
We get the best score for TransE on the WN11 validation set when using $\mathit{l}_1$-norm, learning rate at 0.01, margin $\gamma$ = 6 and d = 50.

In preliminary experiments, we see the highest accuracies on the validation sets for both datasets when using a single memory slot (i.e., $N=1$); and this is consistent with utilizing the single memory slot in language modeling \citep{santoro2018relational}. 
Therefore, we set $N=1$ to use the single memory slot for the triple classification task.
Also from preliminary experiments, we select the batch size $bs=16$ for WN11 and $bs=256$ for FB13, and set the window size $m$ of filters to 1 (i.e., $m=1$).

Regarding other hyper-parameters, we vary the number of attention heads $H$ in \{1, 2, 3\}, the head size $n$ in \{128, 256, 512, 1024\}, the number of MLP layers $l$ in \{2, 3, 4\}, and the number of filters $F=|\bold{\Omega}|$ in \{128, 256, 512, 1024\}.
The memory size $k$ is set to be $nH=k$.
To learn our model parameters, we train our model using the Adam initial learning rate $lr$ in $\{1e^{-6}, 5e^{-6}, 1e^{-5}, 5e^{-5}, 1e^{-4}, 5e^{-4}\}$.
We run up to 30 epochs and use a grid search to select the optimal hyper-parameters.
We monitor the accuracy after each training epoch to compute the relation-specific threshold $\theta_r$ to get the optimal hyper-parameters (w.r.t the highest accuracy) on the validation set, and to report the final accuracy on the test set.

\subsubsection{Search personalization}
We use the same initialization of user profile, query and document embeddings used by \citet{Nguyen2019CapsE} on SEARCH17 to initialize the corresponding embeddings in our R-MeN respectively. 
From the preliminary experiments, we set $N=1$, $bs=16$ and $m=1$. 
Other hyper-parameters are varied as same as used in the triple classification task. 
We monitor the MRR score after each training epoch to obtain the highest MRR score on the validation set to report the final scores on the test set. 

\section{Main results}
\subsection{Triple classification}

\begin{table}[!ht]
\centering
\resizebox{7.75cm}{!}{
\def\arraystretch{1.1}
\begin{tabular}{l|cc|c}
\hline
\bf Method &\bf WN11 & \bf FB13 & \bf Avg.  \\
\hline
NTN \citep{NIPS2013_5028} & 86.2 & 87.2 & 86.7 \\
TransH \citep{AAAI148531} & 78.8 & 83.3 & 81.1 \\
TransR \citep{AAAI159571} & 85.9 & 82.5 & 84.2\\
TransD \citep{ji-EtAl:2015:ACL-IJCNLP} & {86.4} & \textbf{89.1} & {87.8} \\ 
TransR-FT \citep{FengHWZHZ16} & 86.6 & 82.9 & 84.8 \\
TranSparse-S \citep{JiLH016} & {86.4} & 88.2 & 87.3\\
TranSparse-US \citep{JiLH016} & {86.8} & 87.5 & 87.2\\
ManifoldE \citep{ijcai/0005HZ16} & {87.5} & 87.2 & 87.4 \\
TransG \citep{xiao-huang-zhu:2016:P16-1} & 87.4 & 87.3 & 87.4 \\
lppTransD \citep{yoon-EtAl:2016:N16-1}  & 86.2 & {88.6} & 87.4 \\
ConvKB \citep{Nguyen2018ConvKBfull} & {87.6} & {88.8} & {88.2}\\
TransE \citep{NIPS2013_5071} (ours) & \underline{89.2} & {88.1} &  \underline{88.7}\\
\hline
Our R-MeN model & \textbf{90.5} & \underline{88.9} & \textbf{89.7}\\
\hline
\hline
TransE-NMM \citep{Nguyen2016} & {86.8} & 88.6 & 87.7\\
TEKE\_H  \citep{WangL16}  & 84.8 & 84.2 & 84.5\\
Bilinear-\textsc{comp} \citep{guu2015traversing} & 87.6 & 86.1 & 86.9 \\
TransE-\textsc{comp} \citep{guu2015traversing} & 84.9 & 87.6 & 86.3 \\
\hline
\end{tabular}
}
\caption{Accuracy results (in \%) on the WN11 and FB13 test sets. The last 4 rows report accuracies of the models that use relation paths or incorporate with a large external corpus. The best score is in bold while the second best score is in underline. ``Avg.'' denotes the averaged accuracy over two datasets.}
\label{tab:triplecls}
\end{table}

Table \ref{tab:triplecls} reports the accuracy results of our R-MeN model and previously published results on WN11 and FB13.
R-MeN sets a new state-of-the-art accuracy of 90.5\% that significantly outperforms other models on WN11. 
R-MeN also achieves a second highest accuracy of 88.9\% on FB13.
Overall, R-MeN yields the best performance averaged over these two datasets.

\begin{figure}[!ht]
\centering
\includegraphics[width=0.475\textwidth]{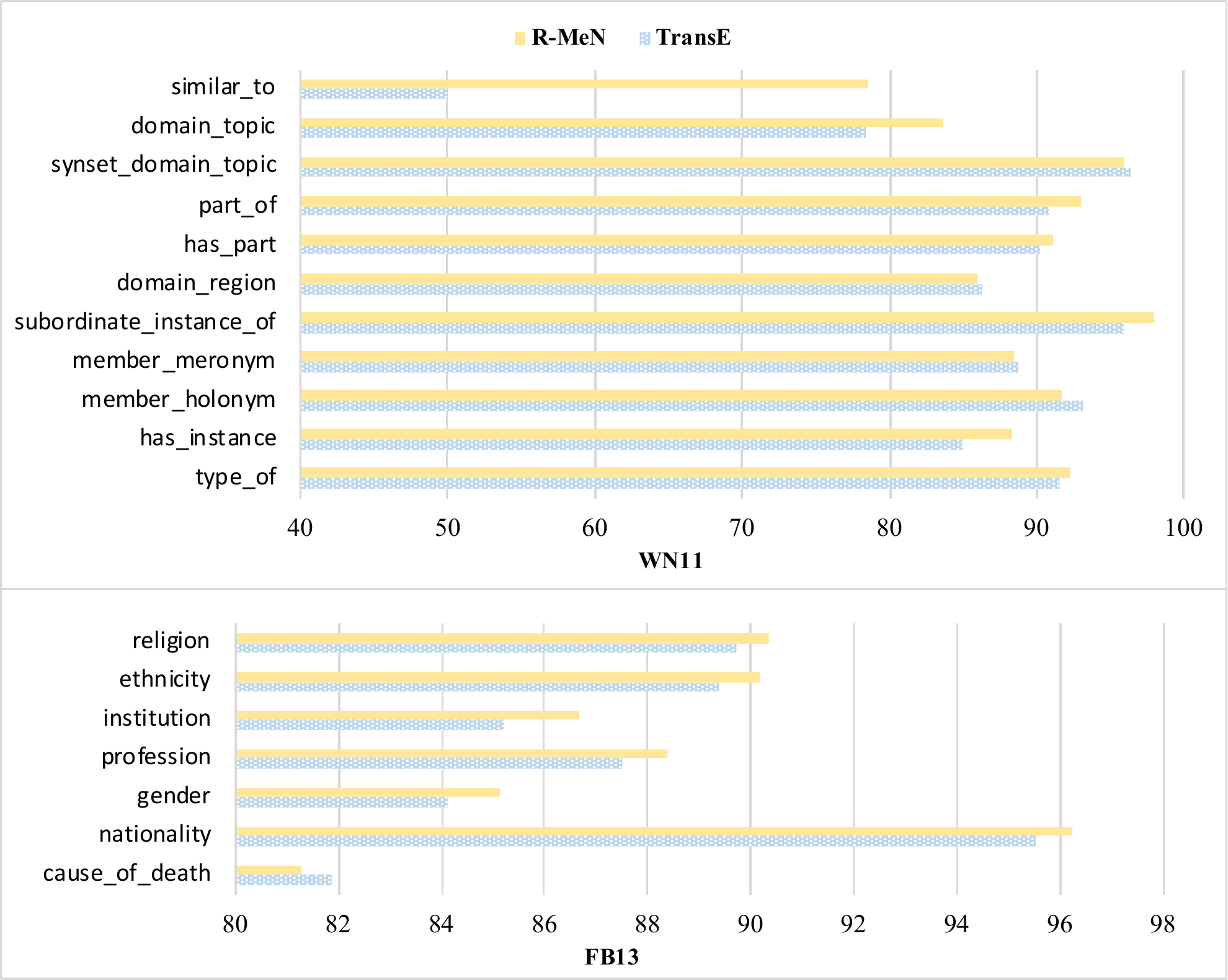}
\caption{Accuracies for R-MeN and TransE w.r.t each relation on WN11 and FB13.}
\label{fig:relationalaccuracies}
\end{figure}

Regarding TransE, we obtain the second-best accuracy of 89.2\% on WN11 and a competitive accuracy of 88.1\% on FB13. 
Figure \ref{fig:relationalaccuracies} shows the accuracy results for TransE and our R-MeN w.r.t each relation. 
In particular, on WN11, the accuracy for the one-to-one relation ``similar\_to'' significantly increases from 50.0\% for TransE to 78.6\% for R-MeN. 
On FB13, R-MeN improves the accuracies over TransE for the many-to-many relations ``institution'' and ``profession''.

\subsection{Search personalization}

\begin{table}[!ht]
\centering
\resizebox{7cm}{!}{
\def\arraystretch{1.1}
\begin{tabular}{l|ll}
\hline
\textbf{Method} & \textbf{MRR} & \textbf{H@1}\\
\hline
SE (Original rank) & 0.559 & 38.5 \\
CI \citep{Teevan2011} & 0.597 & 41.6 \\
SP \citep{Vu2015} & 0.631 & 45.2 \\
\hline
TransE \citep{NIPS2013_5071} & 0.669 & 50.9 \\
ConvKB \citep{Nguyen2018ConvKBfull} & 0.750 & 59.9 \\
CapsE \citep{Nguyen2019CapsE} & 0.766 & 62.1 \\
\hline
Our \textbf{R-MeN} & \textbf{0.778} & \textbf{63.6} \\
\hline
\end{tabular}
}
\caption{Experimental results on the SEARCH17 test set. Hits@1 (H@1) is reported in \%. Our improvements over all baselines are statistically significant with p $<$ 0.05 using the paired t-test.}
\label{tab:resultssp}
\end{table}

Table \ref{tab:resultssp} presents the experimental results on SEARCH17, where R-MeN outperforms up-to-date embedding models and obtains the new highest performances for both MRR and Hits@1 metrics. 
We restate the prospective strategy proposed by \citet{vu2017search} in utilizing the KG embedding methods to improve the ranking quality of the personalized search systems.

\subsection{Effects of hyper-parameters}

\begin{table}[!h]
\centering
\includegraphics[width=0.235\textwidth]{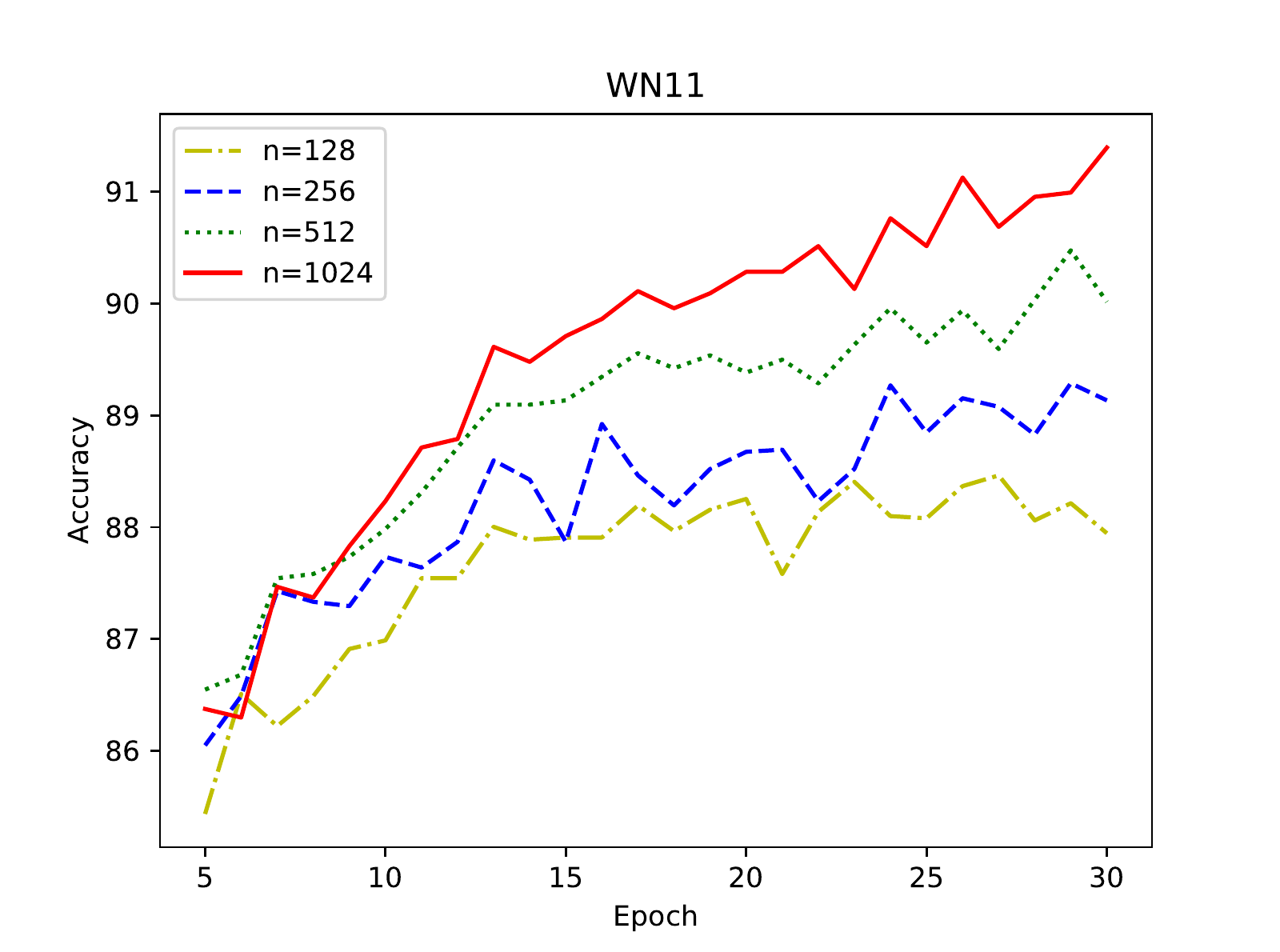}
\includegraphics[width=0.235\textwidth]{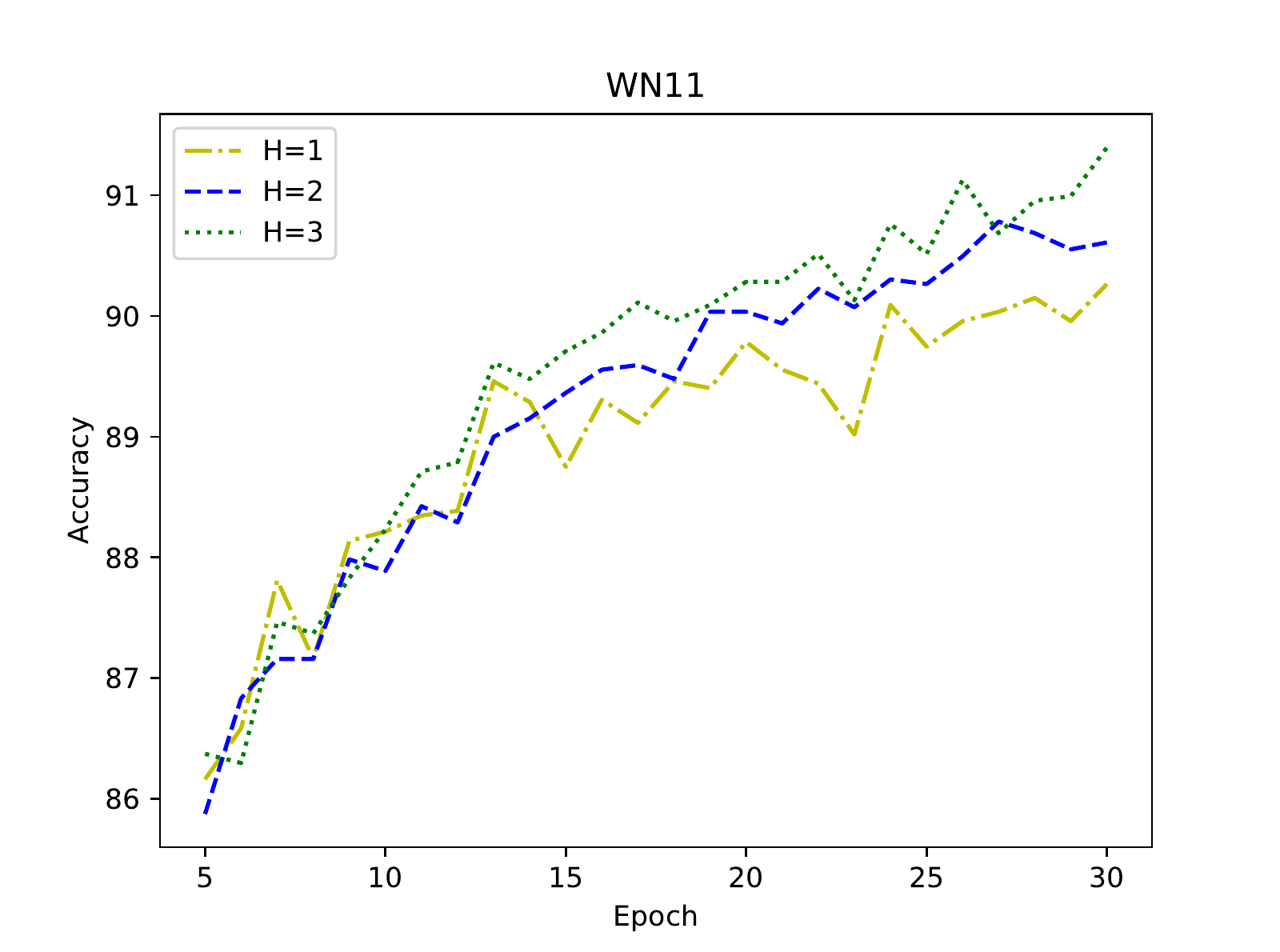}
\includegraphics[width=0.235\textwidth]{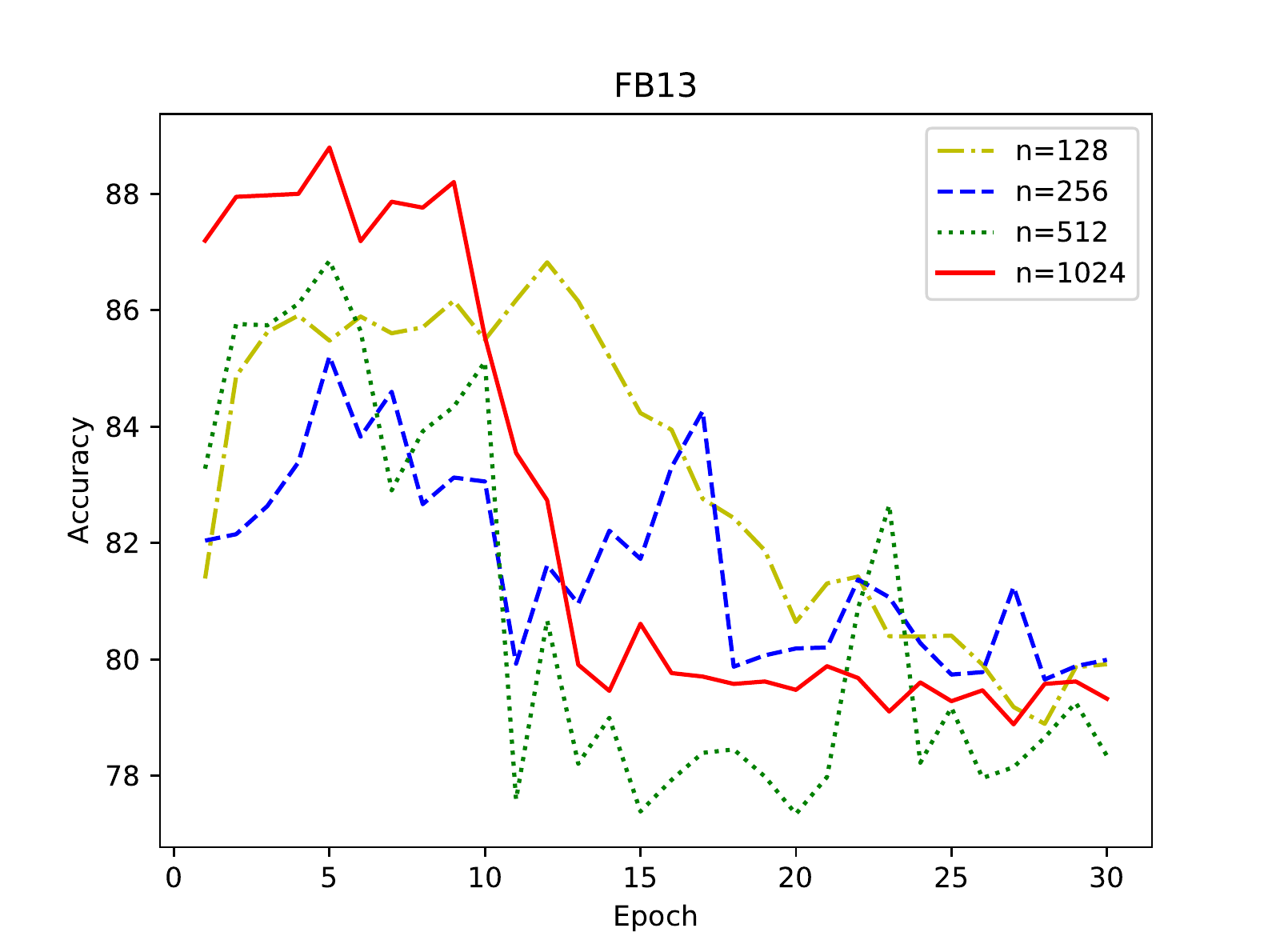}
\includegraphics[width=0.235\textwidth]{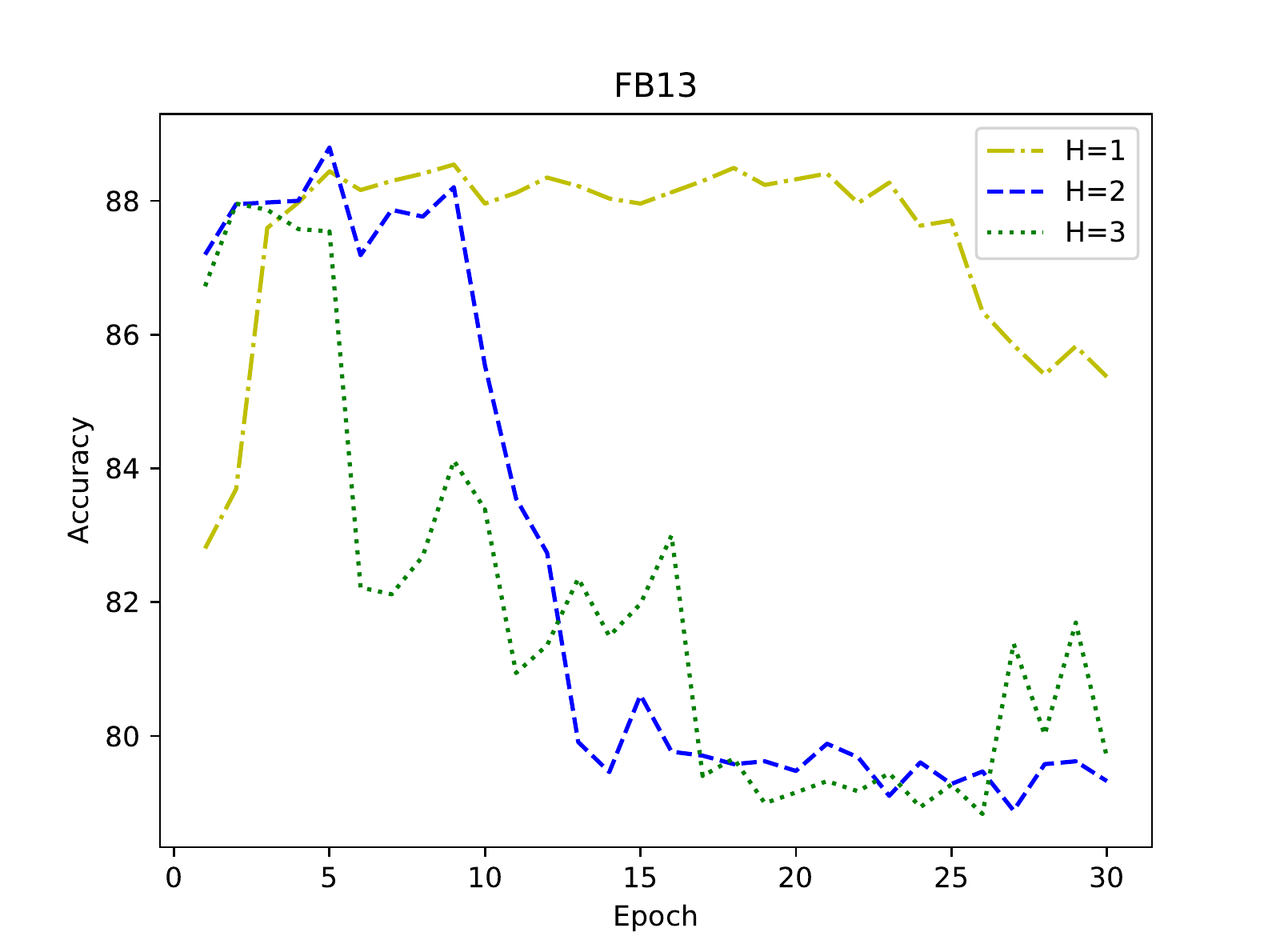}
\includegraphics[width=0.235\textwidth]{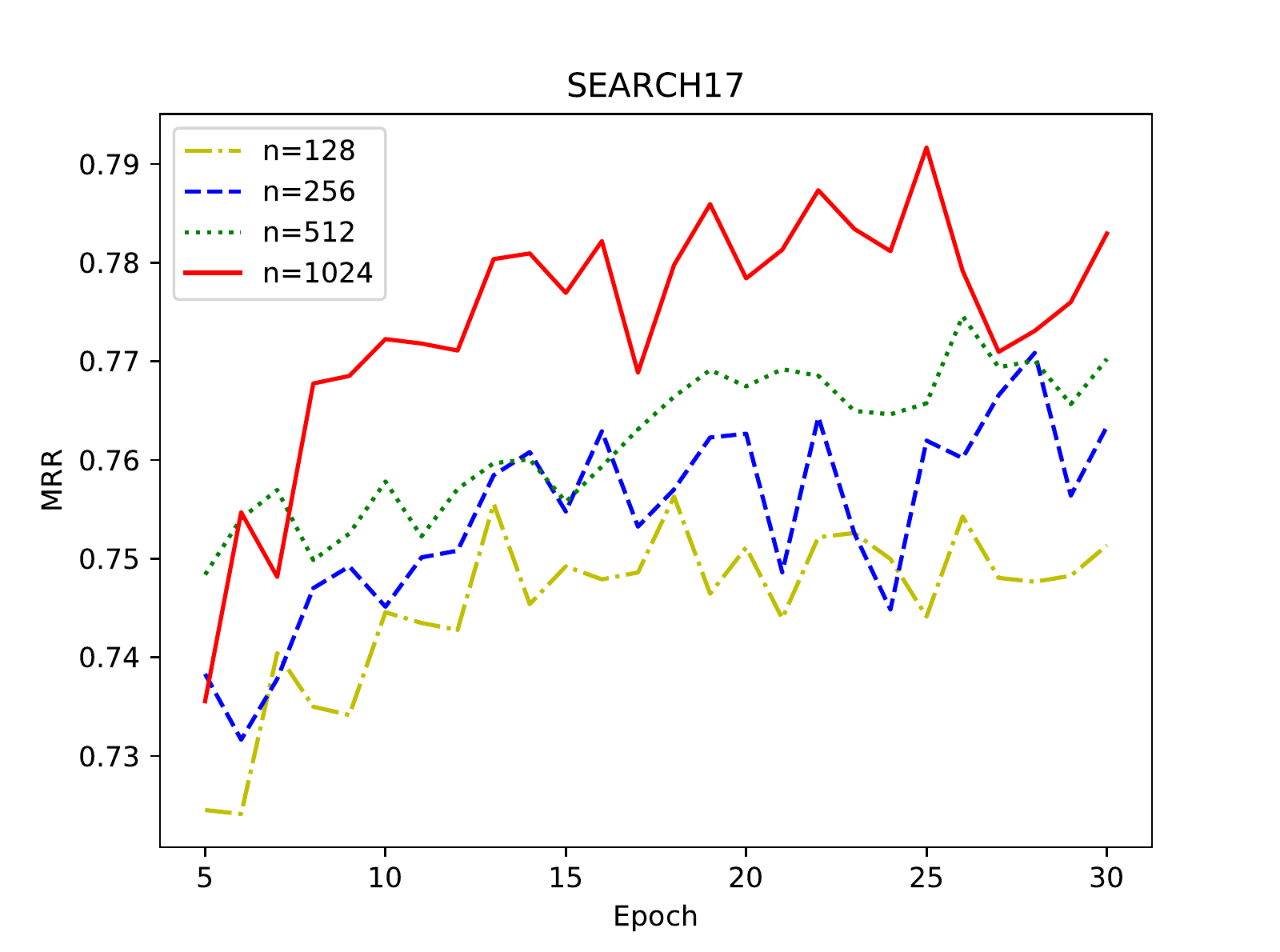}
\includegraphics[width=0.235\textwidth]{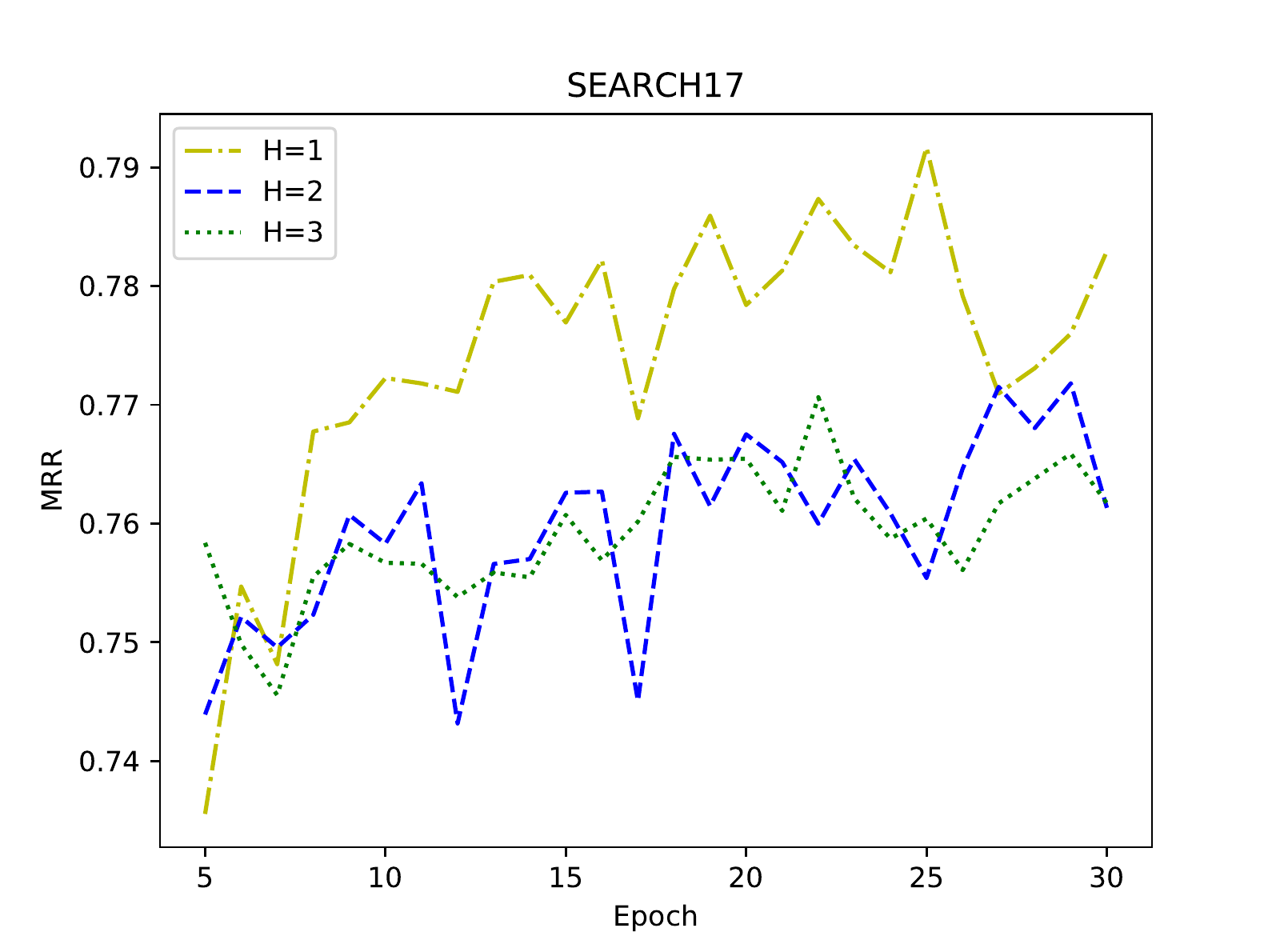}
\captionof{figure}{Effects of the head size $n$ and the number $H$ of attention heads on the validation sets.}
\label{fig:effetsFn}
\end{table}

Next, we present in Figure \ref{fig:effetsFn} the effects of hyper-parameters consisting of the head size $n$, and the number $H$ of attention heads.
Using large head sizes (e.g., $n=1024$) can produce better performances on all 3 datasets. 
Additionally, using multiple heads gives better results on WN11 and FB13, while using a single head (i.e., $H=1$) works best on SEARCH17 because each query usually has a single intention.

\subsection{Ablation analysis}

\begin{table}[!ht]
\centering
\resizebox{7cm}{!}{
\def\arraystretch{1.1}
\begin{tabular}{l|c|c|c}
\hline
\bf Model &  \bf {WN11} & \bf {FB13} & \bf {SEARCH17}\\
\hline
Our \textbf{R-MeN} & \textbf{91.3} & \textbf{88.8} & \textbf{0.792}\\ 
\hdashline
 \ \ \ \ \ (a) w/o Pos & 91.3 & 88.7 & 0.787\\
 \hdashline
 \ \ \ \ \ (b) w/o M & 89.6 & 88.4 & 0.771\\
\hline
\end{tabular}
}
\caption{Ablation results on the {validation} sets. (i) Without using the positional embeddings. (ii) Without using the relational memory network, thus we define $f\left(s,r,o\right) = \mathsf{max}\left(\mathsf{ReLU}\left([\boldsymbol{\mathsf{v}}_s, \boldsymbol{\mathsf{v}}_r, \boldsymbol{\mathsf{v}}_o]\ast\bold{\Omega}\right)\right)^\mathsf{T} \bold{w}$.
}
\label{tab:expresultsDev}
\end{table}

For the last experiment, we compute and report our ablation results over 2 factors in Table \ref{tab:expresultsDev}.
In particular, the scores degrade on FB13 and SEARCH17 when not using the positional embeddings.
More importantly, the results degrade on all 3 datasets without using the relational memory network.
These show that using the positional embeddings can explore the relative positions among $s$, $r$ and $o$; besides, using the relational memory network helps to memorize and encode the potential dependencies among relations and entities.

\section{Conclusion}
\label{sec:conclusion}

We propose a new KG embedding model, named R-MeN, where we integrate transformer self-attention mechanism-based memory interactions with a CNN decoder to capture the potential dependencies in the KG triples effectively.
Experimental results show that our proposed R-MeN obtains the new state-of-the-art performances for both the triple classification and search personalization tasks.
In future work, we plan to extend R-MeN for multi-hop knowledge graph reasoning.
Our code is available at: \url{https://github.com/daiquocnguyen/R-MeN}.

\section*{Acknowledgements}
This research was partially supported by the ARC Discovery Projects DP150100031 and DP160103934.

\bibliography{references}
\bibliographystyle{acl_natbib}

\end{document}